\documentclass[10pt,journal,compsoc]{IEEEtran}
\newif\ifpeerreview
\peerreviewfalse

\usepackage{url}
\usepackage{amsmath,amssymb,graphicx}
\usepackage{xcolor}
\usepackage{color, colortbl}
\usepackage{booktabs}
\usepackage[numbers,sort&compress]{natbib}
\usepackage{lipsum} 

\usepackage[normalem]{ulem} 

\usepackage[switch]{lineno}

\newcommand{\paperID}{64}

\title{Hardware-aware Coding Function Design for Compressive Single-photon 3D Cameras}

\author{David~Parra,Felipe~Gutierrez-Barragan,Trevor~Seets,and~Andreas~Velten
\IEEEcompsocitemizethanks{\IEEEcompsocthanksitem D. Parra, T. Seets, and A. Velten are with the University of Wisconsin-Madison, Madison, WI, 53706. (email deparra@wisc.edu; velten@wisc.edu).\protect\\
\IEEEcompsocthanksitem F. Gutierrez-Barragan is an independent researcher.}
}

\begin{document}

\IEEEtitleabstractindextext{%
\begin{abstract}

Single-photon cameras are becoming increasingly popular in time-of-flight 3D imaging because they can time-tag individual photons with extreme resolution. 
However, their performance is susceptible to hardware limitations, such as system bandwidth, maximum laser power, sensor data rates, and in-sensor memory and compute resources. Compressive histograms were recently introduced as a solution to the challenge of data rates through an online in-sensor compression of photon timestamp data. Although compressive histograms work within limited in-sensor memory and computational resources, they underperform when subjected to real-world illumination hardware constraints. To address this, we present a constrained optimization approach for designing practical coding functions for compressive single-photon 3D imaging. Using gradient descent, we jointly optimize an illumination and coding matrix (i.e., the coding functions) that adheres to hardware constraints. We show through extensive simulations that our coding functions consistently outperform traditional coding designs under both bandwidth and peak power constraints. This advantage is particularly pronounced in systems constrained by peak power. Finally, we show that our approach adapts to arbitrary parameterized impulse responses by evaluating it on a real-world system with a non-ideal impulse response function.

\end{abstract}

}

\maketitle

\IEEEraisesectionheading{
  \section{Introduction}\label{sec:introduction}
}

\IEEEPARstart{S}{ingle-photon} cameras (SPCs) based on single-photon avalanche diodes (SPADs) are capable of measuring the time-of-arrival of individual photons~\cite{spad1, spad2} with extreme time resolution, down to picoseconds~\cite{pellegrini2000laser}. 
This makes SPAD-based SPCs a popular sensor for time-resolved imaging where a scene is illuminated with a fast pulsed periodic light signal, and the SPAD measures the reflected light waveform. These imaging modalities include time-of-flight (ToF) 3D imaging~\cite{tachella2019realtime, Gupta_CVPR2019, rapp2021highflux, rapp2020advances}, non-line-of-sight (NLoS) imaging~\cite{Buttafava2015, liu2019phasor_nlos}, and fluorescence lifetime (FLIM) imaging~\cite{10.1145/3325136}. In particular, single-photon 3D ToF cameras are used in applications such as autonomous vehicles~\cite{OusterOS1} and mobile devices~\cite{zhang2021240}, each with different constraints. As single-photon 3D cameras are adopted in more domains and devices, maximizing their accuracy while adhering to hardware constraints becomes critical.

The illumination and sensor hardware limitations of a single-photon 3D camera play an important role in determining their cost and practicality. 
For instance, although ultra-short pulsed lasers may be a desirable illumination source, they require sophisticated electronics with high bandwidth~\cite{7337356}, which can be impractical and expensive~\cite{10.1063/1.3632117, Upputuri:15}.
Furthermore, thermal, electrical, and material constraints on laser diodes limit the laser signal peak power~\cite{spi_5332, JAOUEN2006163}.
Finally, a precise 3D ToF camera must account for irregular pulse waveforms that may deviate from an assumed model (e.g., Gaussian pulses).
Therefore, bandwidth, peak power, and illumination waveforms are key illumination hardware constraints that should be considered in designing a practical single-photon 3D camera.

\begin{figure}[tp]
    \centering
    \includegraphics[width=\linewidth]{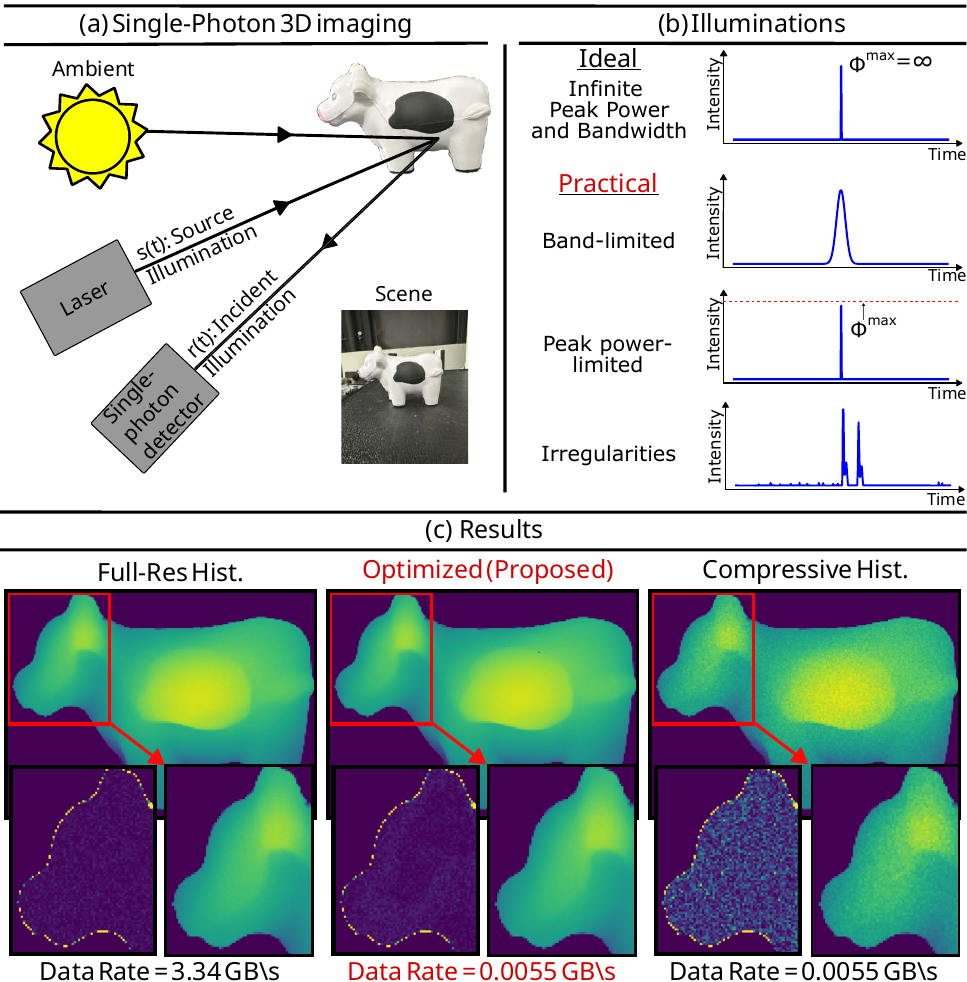}
    \caption{\textbf{Hardware-aware coding function design for compressive single-photon 3D cameras.} (a) Overview of a single-photon 3D camera based on the time-of-flight principle, which illuminates a scene with a pulsed laser and detects the reflected signal at a single-photon detector. (b) The ideal source illumination will have infinite bandwidth and power, but practical hardware considerations demonstrate illumination bandwidth limitations, peak power limitations, and non-ideal impulse response functions. (c) Simulated depth map and depth error results using full-res. histogramming, compressive histogramming~\cite{sheehan2021sketching}, and our optimized coding functions with finite bandwidth and finite peak power.}
    \label{fig:teaser}
\end{figure}

On the sensor side, data rates and the amount of in-sensor compute and memory are constrained.
SPAD pixels in single-photon 3D cameras can generate billions of photon timing events per second.
Although in-sensor timestamp histogramming can make this large volume more manageable~\cite{zhang2021240}, the in-sensor memory required increases proportionally to the temporal and spatial resolution, making it unfeasible to store high-resolution histograms~\cite{gutierrez2022compressive}.
Moreover, regardless of the limited in-sensor memory, transferring high-resolution histograms from a megapixel SPAD array~\cite{ulku2019spad, a115aae9eb73483289b7a22e644bbca0} will lead to data rates of hundreds of Gbit/s~\cite{Gutierrez-Barragan_2023_ICCV} exceeding the limits of the commonly used MIPI CSI-2 implementations ($\sim$10 Gbit/s~\cite{MIPI_CSI2_Brief}).
Hence, to meet data rate constraints, single-photon 3D cameras must perform in-sensor compression of the photon data stream, and this compression algorithm must also function within the sensor’s limited computational and memory resources.

The compressive histograms framework provides a solution for online in-sensor compression of photon timestamp data~\cite{gutierrez2022compressive, sheehan2021sketching, Gutierrez-Barragan_2023_ICCV}.
Compressive histograms encode each photon's timestamp information through a coded projection with multiple ``coding functions"~\cite{gutierrez2022compressive}. 
Recent on-chip implementations demonstrate that compressive histograms can adhere to the highly constrained in-sensor compute and memory resources~\cite{ardelean2023spad, Gutierrez-Barragan_2023_ICCV}.
However, current solutions do not consider illumination hardware constraints (Fig.~\ref{fig:teaser}(b)) in the coding function design, which can lead to sub-optimal performance in real-world systems, as shown in Fig.~\ref{fig:teaser}(c). 
This paper investigates the coding function design for compressive single-photon 3D cameras while accounting for hardware constraints.

To address this problem, we formulate it as a constrained optimization that jointly optimizes the illumination and the corresponding coding matrix, i.e., the coding functions
Given a specific hardware setup, our method identifies a set of high-performing coding functions that adhere to these constraints with minimal loss in depth accuracy. 
These coding functions can adapt to arbitrary parametrized impulse response functions (IRFs), including those affected by irregularities (Fig.~\ref{fig:teaser}(b)). 
Through simulations, we demonstrate that our method achieves superior depth accuracy compared to prior compressive histogram solutions~\cite{gutierrez2022compressive, sheehan2021sketching, gyongy2020highspeed} in hardware-constrained systems. Notably, peak power constraints lead to an optimized illumination with a symmetric, multi-peaked waveform that shows significant improvement over traditional methods (Fig.~\ref{fig:teaser}(c)). Finally, we show that our optimized coding function can adapt to real-world systems with a non-ideal impulse response function and evaluate its performance.

\section{Related Works}\label{sec:related}
\noindent \textbf{Compression in single-photon imaging:} One way to reduce data rates is compressive histogramming~\cite{sheehan2021sketching, gutierrez2022compressive, Gutierrez-Barragan_2023_ICCV, poisson2022luminance}, which is an online, in-sensor compression of histogram data. A simple approach to compressive histograms is coarse histograms~\cite{10.1063/1.3632117, Ren:18}. Although several studies have made an effort to improve coarse histograms~\cite{gyongy2020highspeed, rapp2020advances, 8595429}, it has been shown to perform sub-optimally compared to approaches based on Fourier~\cite{gutierrez2022compressive, sheehan2021sketching, Sheehan2021} and Gray~\cite{gutierrez2022compressive} codes. Recent work has extended compressive histogramming by incorporating spatial and temporal information~\cite{Gutierrez-Barragan_2023_ICCV}. 
Although current compressive histogram methods can account for bandwidth during the depth estimation step~\cite{gutierrez2022compressive}, they fail to use practical illumination constraints in the code design. 
We demonstrate that these codes become suboptimal when applied to real-world systems with band-limited and peak power-limited illuminations. 

Additional strategies to reduce histogram data have been proposed, including equi-depth histograms~\cite{ingle2023count, Sadekar2024EquiDepth} and foveated SPADs~\cite{folden2024foveaspad}. 
Event-based detection has been explored for both passive~\cite{Sundar_2023_ICCV, Sundar_2024_CVPR} and active single-photon imaging~\cite{Yao:24}, which, as a byproduct, reduces data volume by capturing a sparse set of photon data. 
Furthermore, low-resolution ``proximity'' SPAD sensors, which have low data rates, combined with algorithmic upsampling techniques, are a promising low-cost 3D imaging solution~\cite{mu20243dvisionlowcostsinglephoton, li2022deltardepthestimationlightweight, Ruget:25, Sun2022MultiScale, Mora-Martin:23}, but at the cost of a computationally expensive 3D reconstruction step.
In this work, we limit our analysis to incorporating hardware constraints within the compressive histogram framework.

\noindent \textbf{Code optimization in active imaging:} Code design for active 3D imaging has been explored in the areas of indirect ToF (i-ToF)~\cite{gupta2018optimal, gutierrez2019practical, Li_2022_CVPR, Gupta2015Phasor, Kadambi2013Coded, Tadano_2015_ICCV} and structured light (SL)~\cite{Chen_2020_CVPR, Gupta_2011_CVPR, Gupta_2018_ECCV, mirdehghan2018optimal}. Previous work has focused on optimizing codes for real-world systems in both i-ToF~\cite{gutierrez2019practical} and SL~\cite{mirdehghan2018optimal}. However, to the best of our knowledge, hardware-aware code optimization for single-photon imaging has not yet been explored.

\begin{figure*}[tp]
    \centering
    \includegraphics[width=\linewidth]{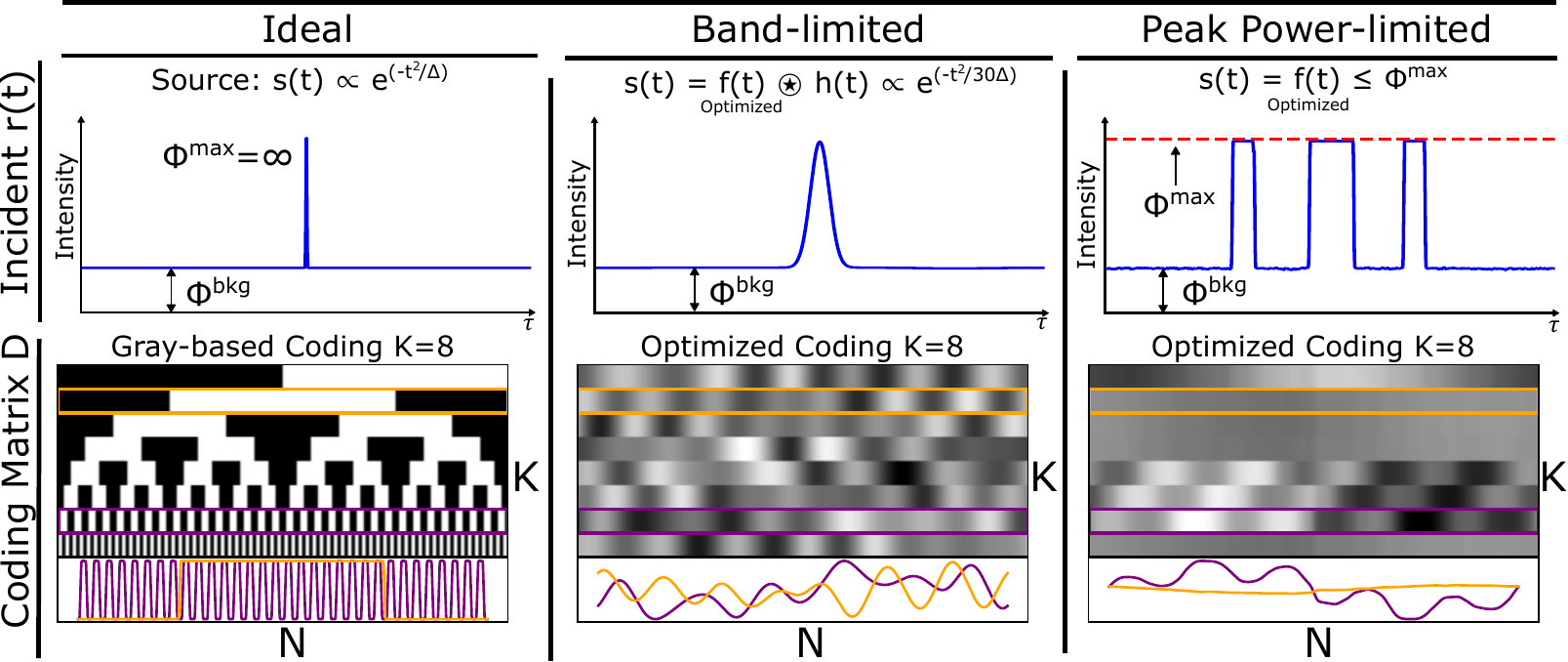}
    \caption{\textbf{Overview of hardware-aware optimized coding functions.} Conventional compressive histograms (left) perform optimally with the ideal pulsed illumination. When hardware constraints, such as bandwidth (middle) and peak power (right), are introduced, compressive histograms become suboptimal, and the coding matrix and source illumination must be adjusted to prevent performance degradation. To address this, we propose a method to optimize both the illumination and coding matrix (or coding functions) based on bandwidth and/or peak power constraints. By adapting the coding functions, we demonstrate improved performance compared to compressive histograms.}
    \label{fig:overview}
\end{figure*}

\section{Single-photon Time of Flight Imaging}\label{sec:background}
Single-photon ToF imaging involves a periodic pulsed laser (of period $\tau$) that illuminates the scene, and a SPAD detector that captures the returned photons. The intensity of the laser at time $t$ can be modeled as a continuous function $s(t) \geq 0$ and the radiance \textit{incident} on the sensor as $r(t) \geq 0$. In practice, the illumination emitted by our system is constrained by its impulse response, defined by the function $h(t)$.

\subsection{Single-photon Histogram Formation}
A single-photon 3D imaging system will convert the incident waveform $r(t)$ into discrete time bins. We denote the number of time bins as $N$, the left edge of the $i$th time bin as $b_i$, and the size of a single time bin as $\Delta$. The number of photons incident on the SPAD follows a Poisson distribution, and the mean is given by,
\begin{equation}
r_i = \Phi_{i}^{\mathrm{sig}} \int_{b_{i}}^{b_{i}+\Delta} s(t -2d/c) dt + \Phi_{i}^{\mathrm{bkg}},
\label{eq:discrete}
\end{equation}
where $d$ is scene distance and $c$ is the speed of light. $\Phi_{i}^{sig}$ is the mean number of signal photons for the $i$th bin, and $\Phi_{i}^{bkg}$ is the background and dark count photon count for the $i$th bin. 

To estimate the incident waveform, a SPAD pixel will form a histogram of photon arrival times. The SPAD pixel starts acquiring photons immediately once triggered by a laser pulse. In each laser cycle, after the first incident photon is detected, the SPAD pixel enters a dead time during which it cannot detect any more photons. The probability $q_i$ that at least one photon is incident during bin $i$ is given by,
\begin{equation}
q_i = 1-e^{\text{-}r_i}.
\end{equation}
A histogram of photon arrival times is constructed over $L$ laser cycles. Let $M_i$ be the number of photons detected in bin $i$. Thus, $M_i$ follows a binomial distribution parameterized by the number of laser cycles $L$ and probability $q_i$. Here, we assume that the SPAD is operated in asynchronous mode~\cite{gupta2019asynchronous} or in the linear regime~\cite {Gupta_CVPR2019} such that pile-up distortion is minimized~\cite {Gupta_CVPR2019}. Thus, the detected histogram is an unbiased estimate of the incident waveform. 

\subsection{Compressive Histograms}

As SPAD ToF cameras increase spatial and temporal resolution, data rates can become unmanageable, making it crucial to find strategies for reducing this. Compressive histograms address this by defining a $K \times N$ \textit{coding matrix} $D$ which will serve as a linear projection on a full-resolution histogram. Thus, the \textit{coded values} are expressed as a matrix multiplication between the measured histogram $M$ and coding matrix $D$, expressed as:
\begin{equation}
B_k = \sum_{i=0}^{N-1} D_{k, i} M_i \text{.}
\label{eq:intent}
\end{equation}
The vector $B=(B_1, ..., B_K)$ represents the coded values for a single pixel. To decode depths, we compute the zero-mean normalized cross-correlations (ZNCC), introduced in SL~\cite{mirdehghan2018optimal}, between our coded values and coding matrix. The decoded depth with respect to time is given as:
\begin{equation}
{\hat{t}}_d \propto \arg\max_{i} \frac{D_{:,i}' - \text{mean}(D_{:,i}')}{||D_{:,i}' - \text{mean}(D_{:,i}')||} \cdot \frac{ (B - \text{mean}(B))} { ||(B - \text{mean}(B))||},
\label{eq:zncc}
\end{equation}
where $D_{:,i}'$ is the $i$th column of $D$ correlated with $h(t)$. ZNCC decoding is a template-matching method that identifies the phase-shifted illumination that best matches the coded values $B$ measured by our SPAD detector. To decode depths for full-res. histograms we use matched filtering~\cite{turin1960}. Since~(\ref{eq:intent}) is a linear operation, compressive histograms can employ an on-the-fly encoding of individual photon timestamps in-sensor, allowing the SPAD pixel to read out the vector $B$ instead of the entire histogram.

\begin{figure}[tp]
    \centering
    \includegraphics[width=\linewidth]{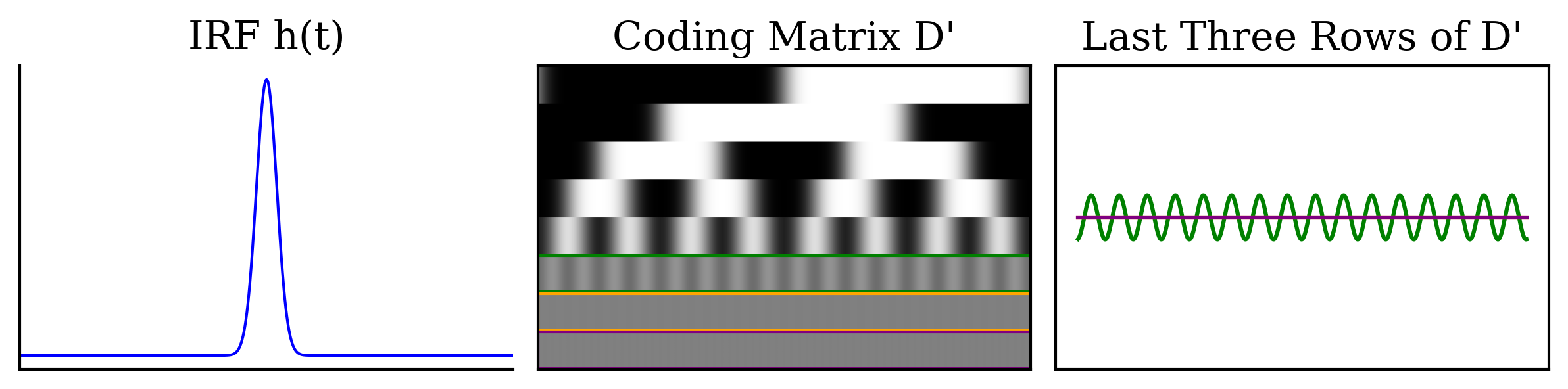}
    \caption{\textbf{Gray codes subjected to strict bandwidth constraints.} Gray-based~\cite{gutierrez2022compressive} compressive histograms cannot efficiently encode the illumination's waveform when severely band-limited.
    }
    \label{fig:toy}
\end{figure}

The compressive histogram framework is the most effective with an ideal pulsed illumination, one that matches the SPAD sensor's temporal resolution and delivers ultra-high peak power. Although this framework has been evaluated using realistic instrument impulse responses, we demonstrate that under realistic hardware conditions, existing compressive histogram codes can perform sub-optimally. As shown in Fig.~\ref{fig:toy}, when using the coding matrix from~\cite{gutierrez2022compressive} with a band-limited illumination, several rows of the matrix $D'$ are zeroed out, causing the corresponding entries in the coded vector $B$ to be zero and contributing nothing to the ZNCC reconstruction. This highlights the need for hardware-aware code design for low-cost, compressive single-photon 3D cameras, which is the focus of this paper.

\section{Hardware-aware coding function design}\label{sec:method}
In this section, we outline the hardware constraints we considered and formulate the constrained optimization for practical single-photon code design.

\subsection{Hardware Limitations}\label{sec:hard}
When developing a low-cost, constrained single-photon 3D camera, it is essential to address several key illumination hardware limitations that will directly impact system performance. While high-end systems may circumvent hardware limitations, low-cost implementations must consider the following constraints:

\smallskip
\noindent\textbf{System Bandwidth:} The bandwidth of our system is characterized by its impulse/frequency response. Ideally, a ToF camera's frequency response should be high enough so that the laser's pulse width matches the timing resolution of our SPAD sensor~\cite{hagen2007gaussian}, assuming sufficiently high photon counts~\cite{shin2016photon}. However, this is not always practical in low-cost, constrained systems. Most LiDAR systems use semiconductor pulsed lasers~\cite{li2024evolution}, which are pulsed by modulating a power supply current. Higher modulation speeds become increasingly ineffective at larger currents~\cite{wen2019large, electronics10070823}, ultimately restricting the bandwidth of the system (about 1-10 ns pulse widths for LiDAR~\cite{Zhai:25}). Consequently, a lower frequency response in our illumination can negatively affect timing resolution~\cite{gutierrez2022compressive} and subsequently depth recovery performance.

\smallskip
\noindent \textbf{Peak laser power:} In addition to matching sensor resolution, the ideal pulsed laser can produce ultra-high peak power. However, the peak intensity achievable in semiconductor lasers is tightly limited due to the highly nonlinear behavior caused by the high gain realized in a small device~\cite{huang1993gain}. High depletion creates intensity-dependent beam distortion and self-focusing that destroys the diode~\cite{kitatani1997room}. Although higher power is achievable with larger diodes, this exacerbates the bandwidth problem. Moreover, due to eye-safety regulations, managing high peak power is crucial, as excessive peak power levels pose a significant risk of eye injury~\cite{ansi-z136.1}. Reported peak optical powers range from $\leq 1$W in short-range flash-LiDARs~\cite{8595429} to hundreds or thousands of watts in long-range scanning systems~\cite{app9194093}. However, reducing peak laser power to meet tighter system limits can cause a decrease in system performance, especially in low signal-to-noise ratio (SNR) regimes.

\smallskip
\noindent\textbf{Impulse response irregularities: } Beyond the limitations of bandwidth and peak power, real-world systems can introduce undesirable characteristics in the IRF, causing it to deviate from well-defined functions like a Gaussian pulse. These irregularities can arise from factors such as misaligned hardware, lens aberrations, and clock jitter. Consequently, these non-ideal impulse responses lead to suboptimal performance of existing compressive histograms in practical single-photon 3D imaging.

\subsection{Constrained Optimization:}
We aim to jointly optimize the illumination and coding matrix under hardware constraints described in section~\ref{sec:hard}. The system's bandwidth limitation is characterized by its impulse response $h(t)$, which acts as a low-pass filter that smooths the input illumination and attenuates frequencies outside the supported bandwidth. Thus, $s(t)$ is the output illumination after the input illumination $f(t)$ is smoothed by $h(t)$. The goal is to design the input illumination $f(t)$. For our simulation results, we model the IRF as a Gaussian pulse with width $\sigma$, i.e., $h(t) \propto e^{-t^2/\sigma}$. However, in real-world systems, $h(t)$ may deviate from a Gaussian shape. Nonetheless, our optimization framework is flexible and can accommodate arbitrary parameterized IRFs. For a pulsed illumination, the output illumination is simply the system's IRF, i.e., $s(t) = h(t)$. The peak power of our system constrains the input illumination, ensuring that $f(t)$ is bounded by the maximum laser power. 
Let $f, h, s$ be vectors representing the $ N$-point discretization of $f(t), h(t), s(t)$, respectively. Therefore, finding $f, D$ can be formulated as a constrained optimization problem, represented mathematically as:

\begin{equation}
\begin{aligned}
\min_{f, D} \quad & l((t_{d_j})_{j=0}^{J-1}, (\hat{t}_{d_{j}})_{j=0}^{J-1}) \\
\text{subject to} \quad 
& (f \circledast h)_i = s_i \\
& f_i \leq \Phi^{\mathrm{max}}, i = 0,\dots,N-1 \\
\end{aligned}
\label{eq:opt}
\end{equation}

where $l$ is the loss function between $J$ true depths $(t_{d_j})_{j=0}^{J-1}$ and predicted depths $(\hat{t}_{d_{j}})_{j=0}^{J-1}$. $\Phi^{\mathrm{max}}$ is the peak photon count, expressed as $\Phi^{\mathrm{max}} = \mathrm{p^{factor}}* \Phi^{\mathrm{sig}}$, where $\Phi^{\mathrm{sig}} = \sum_{i}^{N-1}\Phi_i^{\mathrm{sig}}$ is the mean signal photon count. Bandwidth and IRF constraints are addressed in the second line of~\ref{eq:opt}, whereas the third line contains the peak power limitations. Predicted depth accuracy is strongly influenced by the output illumination $s(t)$ and the coding matrix $D$. Given that the system response $h(t)$ is fixed, our goal is to determine the optimal input illumination $f(t)$ and coding matrix $D$ that minimizes our loss function $l$.

\section{Implementation and Training}\label{sec:implementation}
To solve the constrained optimization problem, we construct a network that can be divided into two sections: one for the coding matrix and one for the illumination. 

\smallskip
\noindent \textbf{Coding matrix section:} The coding matrix $D$ in our network is defined as a 1D convolutional layer with kernel size $N$ and shape $K \times N$, which is equivalent to a fully connected linear layer. This design ensures the coding matrix can be implemented on-chip. To decode depths, we implement a ZNCC layer as defined in~(\ref{eq:zncc}). However, since the argmax operation in the ZNCC depth decoding is not differentiable, we use softargmax.

\smallskip
\noindent \textbf{Illumination section:} The illumination section enforces all the hardware constraints described in section~\ref{sec:hard}. Any bandwidth or IRF limitations are integrated through a 1D convolutional layer that performs the circular convolution between $h$ and $f$. Peak power constraints are implemented through a clamping layer that clips $f$ at the peak photon count $\Phi^{\mathrm{max}}$. For peak power-limited and band-limited illuminations, we filter with $h$ after clamping to remove high frequencies introduced by clipping. To facilitate training, we initialize our illumination to $f_i = 1.0, i = 0,\dots, N-1$.

\begin{figure}[tp]
    \centering
    \includegraphics[width=\linewidth]{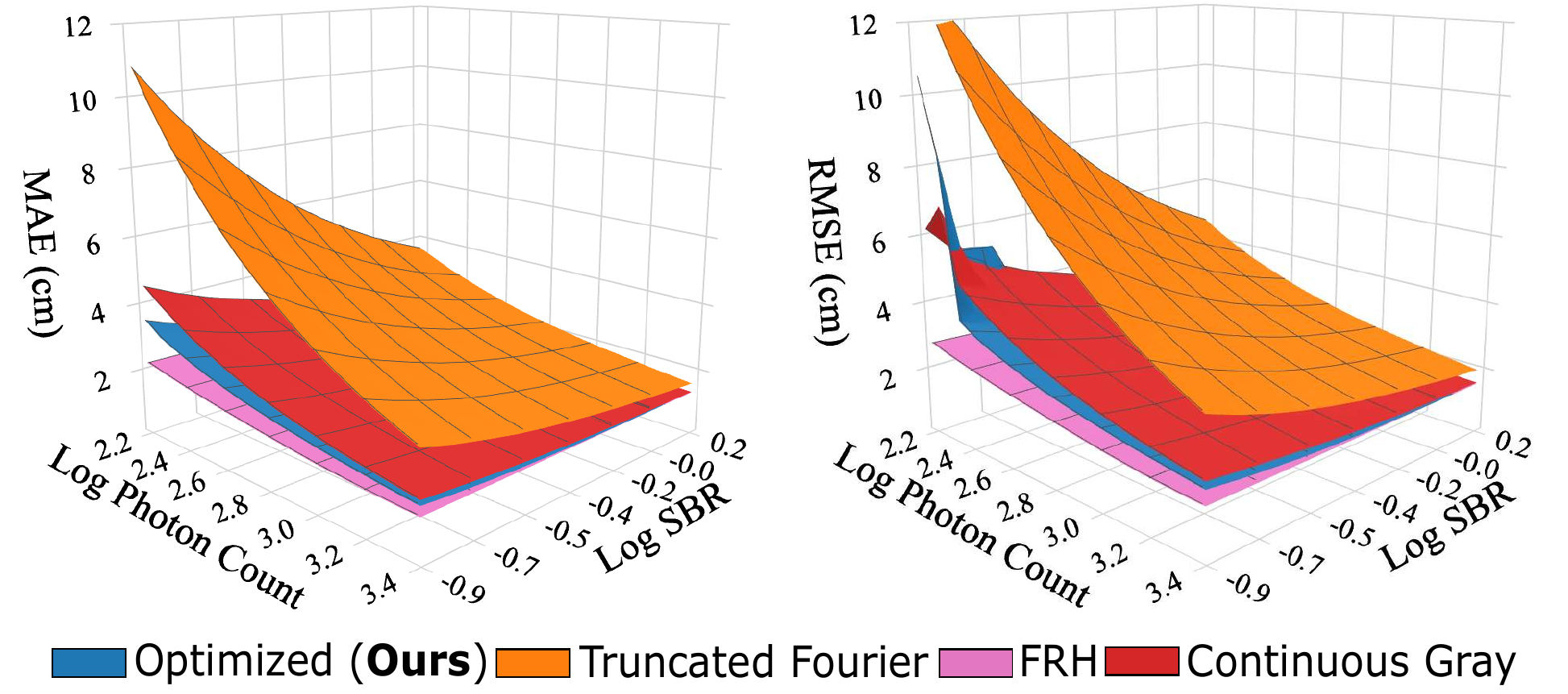}
    \caption{\textbf{Simulated results with finite bandwidth and infinite peak power}. This figure shows the mean absolute error on the left and the root mean squared error on the right, obtained from Monte Carlo simulations using the different coding schemes, evaluated at an IRF width of $\sigma = 30\Delta$ and infinite peak power. The optimized coding functions achieve the closest accuracy to the FRH, regardless of SBR, when the mean photon count is above $150$. For photon counts below $150$, both the Gray and the optimized codings exhibit a large gap between RMSE and MAE due to the presence of outliers.}
    \label{fig:sim_avg}
\end{figure}
\smallskip
\noindent \textbf{Training with Poisson noise:} Since we are optimizing for SPADs, the dominant noise source is Poisson. To simulate a measured histogram, we apply Poisson noise to the incident vector ($(r_i)_{i=0}^{N-1}$) in the network. However, Poisson noise is not differentiable and cannot be implemented without breaking gradient flow. Therefore, to mimic Poisson noise, we approximate it with Gaussian noise where the mean equals the variance. When photon counts are sufficiently high, this Gaussian noise converges to Poisson noise. To ensure our coding functions adapt to SPAD sensors, we test all of our results using Poisson noise.

The network uses gradient descent to jointly optimize both the illumination and the coding matrix. First, the illumination's waveform is optimized according to the provided hardware constraints. Then the coding matrix is adjusted so that each row’s frequency content most effectively samples the optimized illumination. Both the illumination parameters and coding matrix weights are updated together at every backpropagation step.

Note that we generate distinct coding functions corresponding to a specific set of hardware constraints. Although these coding functions vary according to hardware constraints, they remain constant across different SNR levels. Consequently, optimal performance is achieved when the coding functions are applied under the specific hardware constraints for which they were optimized.

\subsection{Training}
The dataset used for training consisted of generated labels, where each label represented a ground truth depth and associated SNR level. For our experiments, coding functions were trained in a moderate SNR regime.  We trained all models using the ADAM optimizer~\cite{kingma2014adam}, and used a learning rate of $0.013$ for bandwidth-limited results and $0.0018$ for peak power-limited results. The learning rate decay was set to $0.3-0.4$ for both the peak power-limited and band-limited results. It took between 10 and 30 epochs to converge to the local minimum, with fewer epochs required for the bandwidth-limited results. We trained using the L1 loss function with total variation (TV) regularization applied to the 1D convolution layer (coding matrix) weights to reduce noise. See the supplement for more training and implementation details.

\section{Evaluation of Optimized Codes}\label{sec:results}
The following section evaluates our optimized coding functions at various SNR levels with respect to hardware limitations discussed in section~\ref{sec:hard}. 

\subsection{Baselines and Performance Metrics}
We compare our method against the following baselines:

\begin{itemize}[]
    \item \textbf{Truncated Fourier~\cite{sheehan2021sketching}:} This coding matrix $D$ is the first $K$ rows of the discrete Fourier transform matrix, not including the zeroth harmonic. 
    \item \textbf{Continuous Gray~\cite{gutierrez2022compressive}}: This coding matrix $D$ is based on the Hamiltonian codes introduced in~\cite{gupta2018optimal}. The rows are the $K$-bit Gray Codes~\cite{gray1953} linearly interpolated to length $N$.
    \item \textbf{Full-resolution Histogramming (FRH):} Coding matrix $D$ is the $N\times N$ identity matrix. Full-resolution histogramming provides no compression but offers the most accurate depth recovery. 
    \label{sec:codes}

\end{itemize}

For the baselines, we use a pulsed illumination. To determine the efficacy of a code, we computed its root mean squared error (RMSE) and mean absolute error (MAE) through Monte Carlo simulations over the depth range for different photon counts and signal-to-background ratio (SBR) levels. RMSE and MAE are easily comparable performance metrics and are well-defined indicators of a code's performance~\cite{gupta2018optimal, gupta2019asynchronous}. See the supplement for further details on performance metrics and Monte Carlo simulation.

\begin{figure*}[tp]
    \centering
    \includegraphics[width=\linewidth]{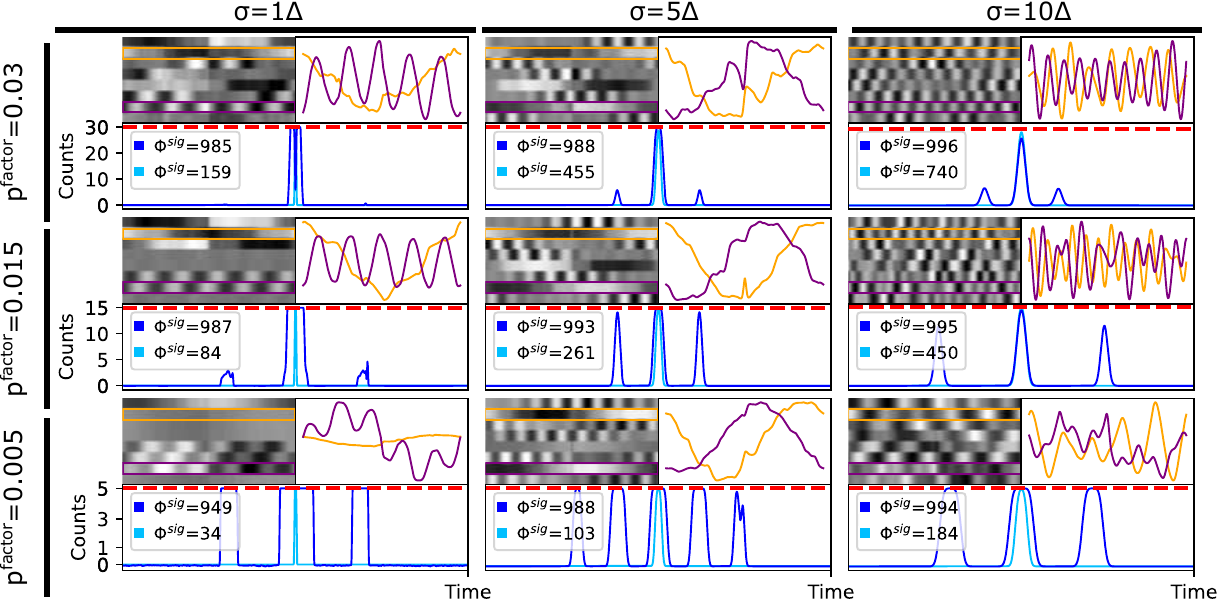}
    \caption{\textbf{Hardware-constrained optimized coding functions.} This figure plots the optimized coding functions found for different bandwidths and peak power constraints. The columns indicate the width of the IRF $h(t)$ and the rows indicate the peak factor $p^{\text{factor}}$. For each $p^{\text{factor}}$ and $\sigma$ we display the coding matrix $D$ on the top right, display the $2$nd and $7$th row of $D$ on the top left, and display the pulsed illumination (light blue) and the optimized illumination (blue) incident on the sensor at the bottom. Here we optimize our codes with a constant mean photon count of $\Phi^{\mathrm{sig}}=1000$. Note that the middle of the illumination plot is zero.}
    \label{fig:peak_illums}
\end{figure*}

\subsection{Optimized Band-limited Coding Functions}
Given finite bandwidth, but infinite peak power, the optimal band-limited illumination converges to the IRF,  $h(t)$. This suggests that, with infinite peak power, the best-performing illumination is the narrowest pulse feasible by the hardware.
Nevertheless, reduced bandwidth leads to an increase in pulse width, making system performance highly dependent on the choice of compressive histogram. Coding matrix rows that contain frequency components beyond the illumination’s bandwidth contribute nothing to reconstruction, as they are zeroed after filtering. High-frequency codes, such as Gray-based codes, will be inefficient at encoding an illumination with a low frequency response. Conversely, codes such as Truncated Fourier are effective for low-frequency waveforms, but may fail to capture higher frequency components as bandwidth increases.  As a result, neither Truncated Fourier nor Gray codes are perfectly suited for varying impulse responses and may underperform with a band-limited illumination. On the other hand, our proposed codes are explicitly optimized to match the system’s impulse response. This leads to a coding matrix that maintains a balance between high and low frequencies and is robust across different bandwidths. See the supplement for visualizations of our optimized band-limited coding functions.

\subsection{Band-limited System Performance}
In this section, we evaluate the performance of the baselines described in Section~\ref{sec:codes} and our optimized coding functions under conditions where the bandwidth is finite but infinite peak laser power. Fig.~\ref{fig:sim_avg} shows the resulting MAE and RMSE from our Monte Carlo simulations with an IRF width of $\sigma = 30\Delta$, where $K = 8$ for our compressive codes and $N = 1024$ for the FRH. As the illumination bandwidth decreases ($\sigma> \Delta$), the effective time resolution of our system decreases, leading to diminished FRH performance compared to the ideal pulse ($\sigma = \Delta$). In this strict bandwidth regime, Gray-based codes experienced the greatest performance degradation, whereas Truncated Fourier exhibited minimal loss. Regardless, we identify an optimized coding function that approaches FRH performance more closely than either compressive histograms. As indicated by the RMSE and MAE, for photon counts above $150$ and SBR above $0.15$, the optimized codes yield the fewest outliers and produce less variance in depth estimation compared to Truncated Fourier and Gray codes. The higher RMSE observed for the optimized coding functions at photon counts $\leq 150$ arises from outliers. We attribute these outliers to training on Gaussian noise (mean equal to variance) as an approximation of Poisson noise, as well as emphasizing medium-to-high photon counts. Nevertheless, the optimized coding functions continue to achieve near-zero errors for the majority of depths, as shown by the MAE Monte Carlo simulation (Fig.~\ref{fig:sim_avg}). See the supplement for additional band-limited results.

\subsection{Optimized Peak Power-limited Coding Functions} Fig.~\ref{fig:peak_illums} shows the resulting illuminations and coding matrices after optimizing for peak factors $\mathrm{p^{factor}}=[0.005, 0.015, 0.030]$, IRF widths $\sigma = [1\Delta, 5\Delta, 10\Delta]$, and a constant signal photon count of $\Phi^{\mathrm{sig}}=1000$. Under conditions of finite bandwidth and peak power, we observe that the optimal illumination diverges from the pulsed waveform. Instead, the optimized illumination converges to a symmetric waveform with multiple peaks, almost resembling a square wave. This behavior becomes more pronounced as the peak photon count decreases. Notably, several of the illuminations are binary and, thus, more efficient to implement in hardware. For conventional pulsed illumination, a peak power limitation can lead to a reduction in the total signal photons  $\Phi^{\text{sig}}$ when clipping. To mitigate this photon loss, the optimized illumination adapts by redistributing photons dependent on the severity of the peak factor $\mathrm{p^{factor}}$. However, as we increase $\mathrm{p^{factor}}$, the optimal illumination converges back to the IRF. Fig.~\ref{fig:peak_illums} illustrates how the coding matrix adapts in response to illumination changes. Notably, for the optimized illumination characterized by an IRF width of $\sigma=1\Delta$ and a peak factor of $\mathrm{p^{factor}}=[0.015,0.005]$, several of the rows within the coding matrix are nearly zero. This implies these near-zero rows do not encode any important frequency information about our illumination. Consequently, only a subset of rows in the coding matrix is necessary for accurate depth decoding, and the remainder can be potentially discarded. Note that the total signal photon count $\Phi^{\mathrm{sig}}$ and peak photon count $\Phi^{\mathrm{max}}$ are fixed during optimization, ensuring that the illumination does not violate these constraints.

\begin{figure}[tp]
    \centering
    \includegraphics[width=\linewidth]{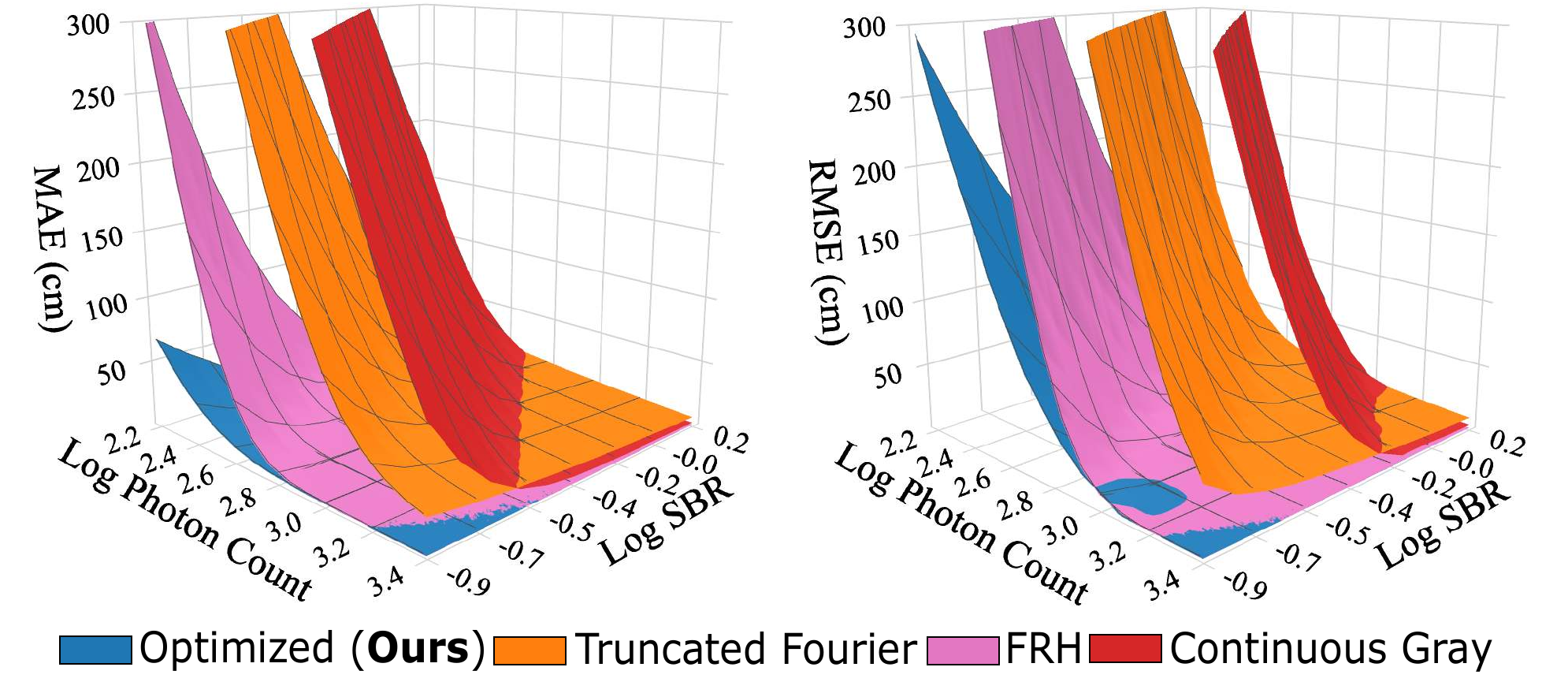}
    \caption{\textbf{Simulated results with finite bandwidth and finite peak power when clipping pulse energy.} This figure shows the MAE on the left and the RMSE on the right, obtained from Monte Carlo simulations using the different coding schemes, evaluated at an IRF width of $\sigma = 5\Delta$ and peak factor of $\mathrm{p^{factor}}=0.05$ (clipped). }
    \label{fig:sim_peak_clip}
\end{figure}

\subsection{Peak Power-Limited System Performance}
In this section, we analyze each of the baselines described in section~\ref{sec:codes} and our optimized coding functions under finite bandwidth and finite peak power. In our experiments, we benchmark the coding functions against pulse-based methods where peak power is constrained in two ways: clipping the maximum pulse energy at $\Phi^{\mathrm{max}}$, and maintaining constant pulse energy by increasing pulse width.

\smallskip
\noindent \textbf{Clipped energy results:} Fig.~\ref{fig:sim_peak_clip} shows the RMSE and MAE for the different compressive codes ($K=8$) and FRH ($N = 1024$) tested with a peak factor of $\mathrm{p^{factor}}=0.005$ and an IRF width of $\sigma = 5\Delta$ when clipping pulse energy at $\Phi^{\mathrm{max}}$.  Clipping significantly degrades the performance of pulse-based methods due to reduced signal photons. In contrast, the optimized illumination adapts by maximizing the number of signal photons permitted by the system while not exceeding the peak photon count. 
Consequently, the optimized coding functions achieve the lowest RMSE and MAE across all photon counts among compressive histogram baselines, and even outperform FRH at counts below $1500$. At low photon counts, pulse-based methods become unstable under clipping, while the optimized coding functions maintain performance despite the limited peak laser power.

\begin{figure}[tp]
    \centering
    \includegraphics[width=\linewidth]{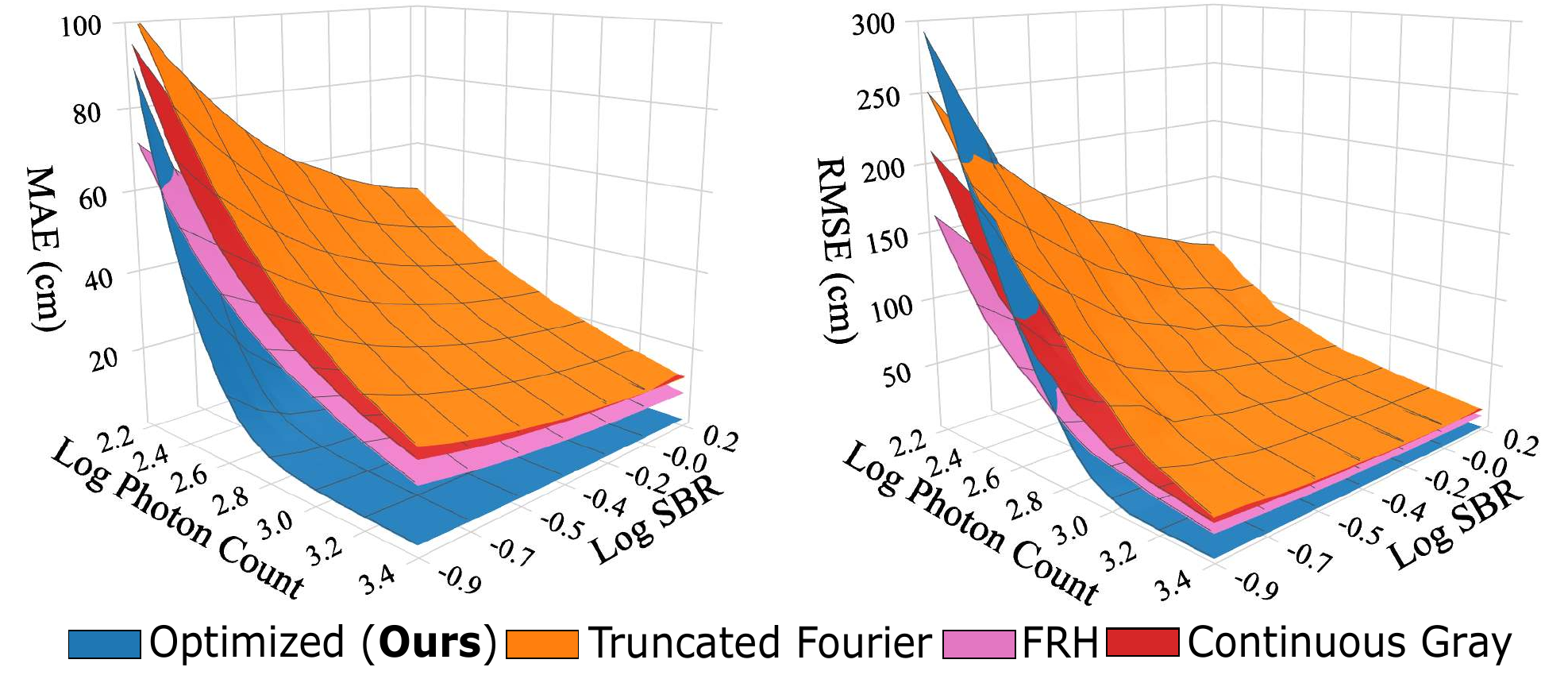}
    \caption{\textbf{Simulated results with finite bandwidth and finite peak power when preserving pulse energy.} This figure shows the MAE on the left and the RMSE on the right, obtained from Monte Carlo simulations using the different coding schemes, evaluated at an IRF width of $\sigma = 5\Delta$ and peak factor of $\mathrm{p^{factor}}=0.05$ (constant). }
    \label{fig:sim_peak_constant}
\end{figure}

\smallskip
\noindent \textbf{Preserving energy results:} Fig.~\ref{fig:sim_peak_constant} shows the RMSE and MAE for the different compressive codes ($K=8$) tested with a peak factor of $\mathrm{p^{factor}}=0.005$ and an IRF width of $\sigma = 5\Delta$ when maintaining constant pulse energy. In this configuration, increasing the pulse width avoids clipping, resulting in significant improvement for pulse-based methods. However, this results in reduced temporal resolution and lower peak power in the pulse waveform. Consequently, preserving pulse energy does not overcome the limitations imposed by peak laser power, which becomes the main bottleneck for the pulsed baselines (at $\mathrm{p^{factor}}=0.005$). As a result, the optimized coding functions demonstrate the lowest RMSE and MAE among compressive histograms and FRH when signal photon counts are above $200$. Notably, FRH never surpasses the optimized codes. For photon counts below $200$, the optimized codes exhibit a wider gap in RMSE and MAE due to occasional outliers (from training on Gaussian noise), yet still achieve near-zero error for the majority of depths.

\smallskip
These findings indicate that clipping and preserving pulse energy yield suboptimal performance of pulse-based methods, even for FRH. To avoid performance loss in low-to-medium peak power regimes, it is essential to reconsider the choice of illumination and coding matrix based on hardware constraints. Please see the supplement for additional peak power-limited results.

\section{Compressive Single-Photon 3D Imaging}
In this section, we evaluate the performance of our proposed coding functions on real-world data from a scanning-based hardware setup~\cite{gupta2019asynchronous}, and simulated flash-illumination scenes.  Additionally, we analyze the memory efficiency of our optimized coding functions for an on-chip implementation.

\subsection{Real-world Scanning-based System Results}
\begin{figure}[tp]
    \centering
    \includegraphics[width=\linewidth]{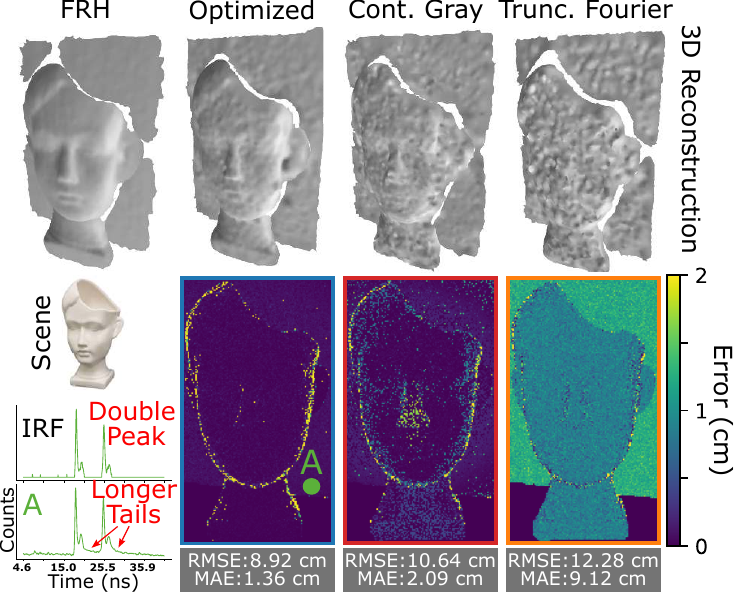}
    \caption{\textbf{3D reconstruction visualization.} Real-world scans of a porcelain face obtained using Gray codes, truncated Fourier, and our optimized codes. RMSE and MAE are calculated from the FRH as the reference. Several background histograms (blue dot) show longer tails, which are likely due to indirect reflections. }
    \label{fig:exp}
\end{figure}

To evaluate the ability of our optimized codes to adapt to real-world hardware, we use data acquired from a scanning-based single-photon system~\cite{gupta2019asynchronous}, which has been preprocessed as described in~\cite{gutierrez2022compressive}. The post-processed histogram data have $\Delta = 8$ ps and $N=2188$. The system IRF, shown in Fig.~\ref{fig:exp}, exhibits a double peak, likely due to lens inter-reflections in the hardware setup. Fig.~\ref{fig:exp} shows the recovered 3D reconstruction of a porcelain face scan from compressive histograms and our optimized coding functions, where $K=10$. When presented with this atypical IRF, compressive histograms suffer from performance degradation, even when the IRF is taken into account. In particular, truncated Fourier exhibits systematic depth errors, implying this code struggles to resolve the double peaks in the IRF. While Gray-based codes provide more stable results, they are prone to moderate depth errors across most pixels and large errors in certain regions (outliers). In contrast, our optimized coding demonstrates robustness to the irregular IRF and yields the closest reconstruction to the FRH. Additionally, the hardware scans in Fig.~\ref{fig:exp} include pixels with dense indirect reflections, which appear as extended tails in the histogram. As a result, all the compressive codes show higher RMSE values due to these pixels. Nonetheless, the optimized coding still recovers most depths with minimal error, while remaining robust to pixels with dense indirect reflections. Note that only the coding matrix was optimized for the hardware results. Please refer to the supplement for additional hardware results.

\subsection{Simulated Flash-Illumination System Results}
To evaluate the performance of our optimized coding functions in a flash-illumination-based system, we used rendered histogram data generated with MitsubaToF~\cite{jakob2010mitsuba, Pediredla2018pileup}, obtained from~\cite{Gutierrez2021iToF2dToF}. Here, we have $N=2048$ for the band-limited results and $N=1024$ for the peak power-limited results. Different illuminations are simulated by convolving the incident waveform with the rendered histograms and then scaling and offsetting the results based on the desired photon count and SBR. The resulting scenes include both dense and sparse indirect reflections.

\begin{figure}[tp]
    \centering
    \includegraphics[width=\linewidth]{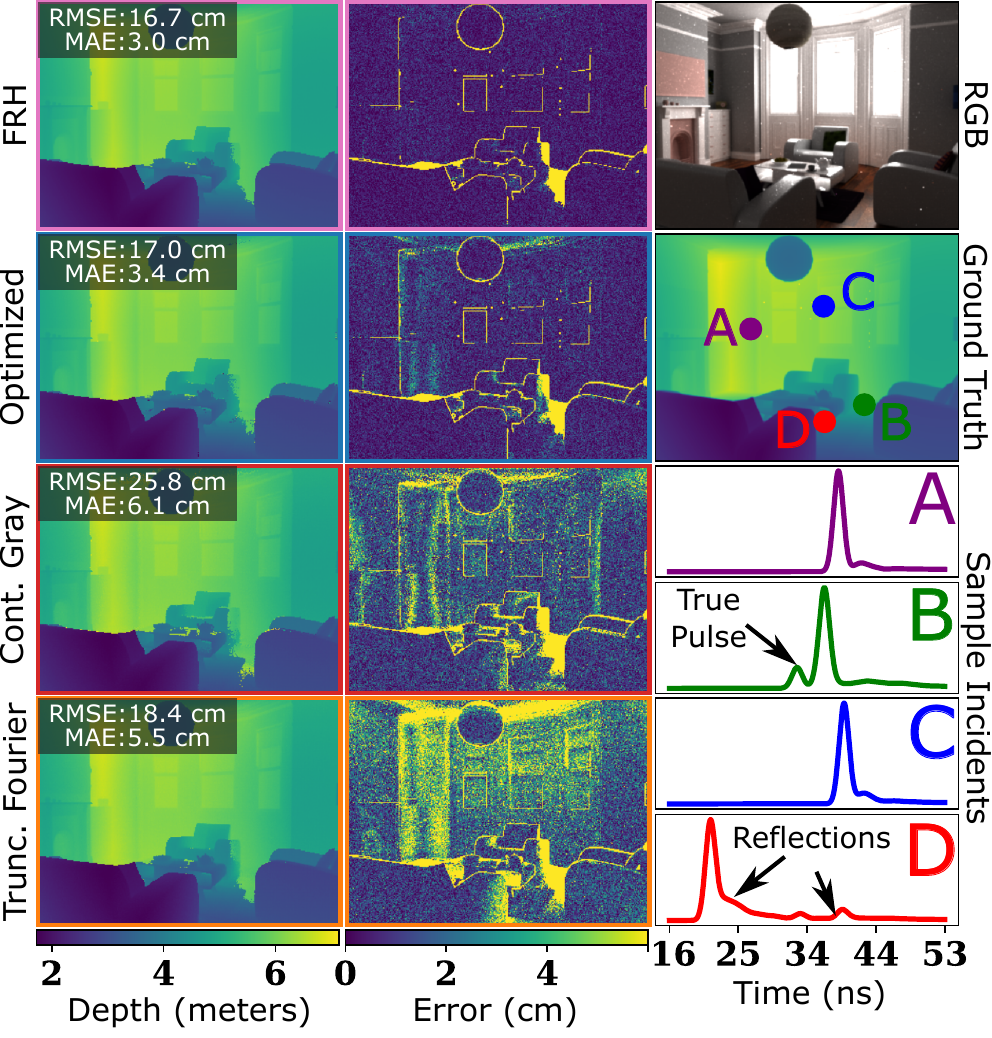}
    \caption{\textbf{Simulated flash-illumination for band-limited coding functions.} Depth maps (first col) and error (second col) results for different coding schemes (rows) evaluated with IRF width $\sigma=30\Delta$ and infinite peak power. The rightmost column plots the RGB image, ground truth depth map, and four incident functions taken from the scene in that order.}
    \label{fig:band-flash-lidar}
\end{figure}

\smallskip
\noindent \textbf{Band-limited coding functions:} Fig.~\ref{fig:band-flash-lidar} plots the resulting depth maps and error for compressive histograms and the optimized coding functions with an IRF width $\sigma=30\Delta$ and infinite peak power where $K=12$ ($K=11$ for Gray codes). Truncated Fourier produces systematic errors in the presence of indirect reflections, but remains relatively robust to large errors. Although Gray-based coding handles indirect reflections more effectively, this advantage diminishes under low-bandwidth constraints. The band-limited optimized coding functions are robust to incident waveforms that do not deviate far from the trained waveform, including both dense and sparse indirect reflections with a small secondary peak. However, they are susceptible to outliers caused by large, sparse indirect reflections. These outliers can be mitigated by increasing $K$ as shown by the figure results. Overall, the optimized coding functions still outperform compressive histograms in terms of both RMSE and MAE, even under moderate indirect reflections.

\begin{figure*}[tp]
    \centering
    \includegraphics[width=\linewidth]{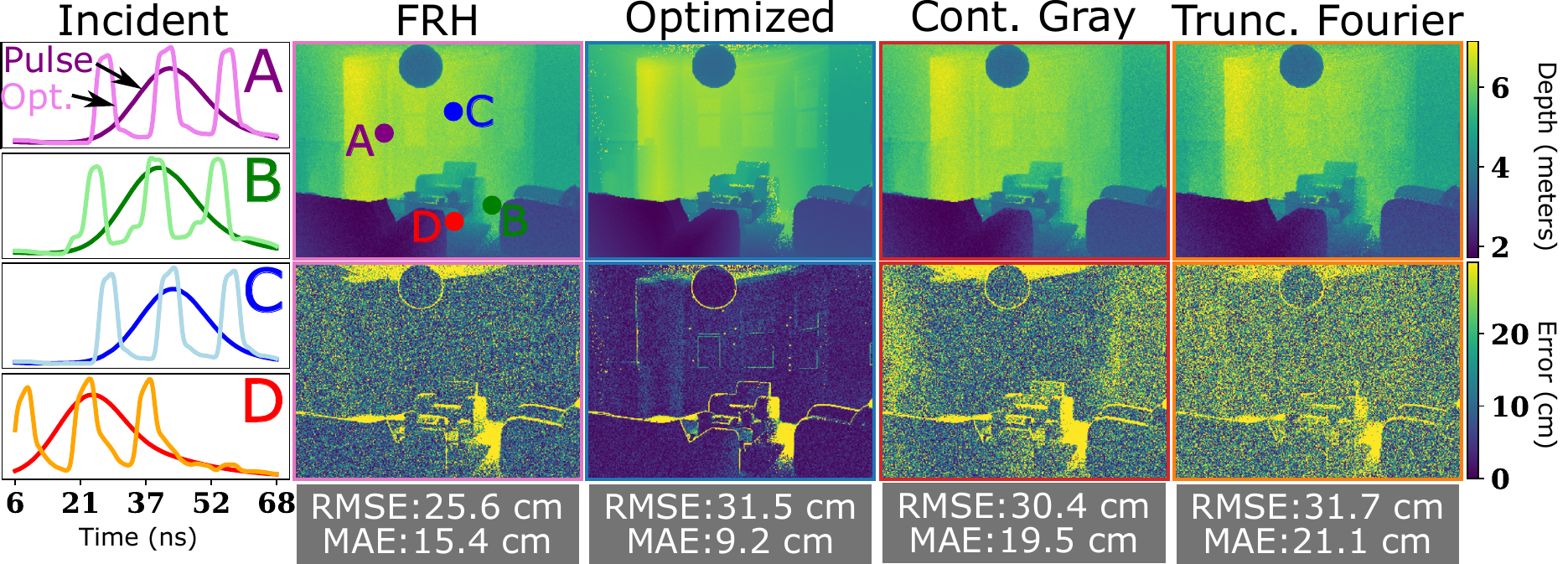}
    \caption{\textbf{Simulated flash-illumination for peak power-limited coding functions.} Depth maps (first row) and error (second row) results for different coding schemes (cols) when preserving pulse energy evaluated at IRF width $\sigma=10\Delta$ and $\mathrm{p^{factor}}=0.005$. The leftmost column plots the four incident functions taken from the scene for both the pulsed and optimized illuminations.}
    \label{fig:peak-flash-lidar}
\end{figure*}

\smallskip
\noindent \textbf{Peak power-limited coding functions:} Fig.~\ref{fig:peak-flash-lidar} plots the resulting depth maps and error for compressive histograms and the optimized coding functions with IRF width $\sigma=10\Delta$ and peak factor of $\mathrm{p^{factor}}=0.005$ when maintaining constant pulse energy. Similar to the band-limited results, the optimized coding functions are prone to outliers in the presence of sparse indirect reflections, which is further exacerbated by the multiple peaks in the peak power-limited illumination. However, in severely constrained systems, peak laser power becomes the limiting factor in coding scheme performance rather than indirect reflections. For the pulse-based methods, preserving energy by widening the pulse does not compensate for the lack of peak laser power.   Consequently, our optimized coding functions achieve the lowest MAE among all baselines (when $\mathrm{p^{factor}}=0.005$), including FRH, despite challenges from indirect reflections. Please refer to the supplement for additional flash-illumination results and an analysis of the optimized coding functions in the presence of different indirect reflections.

\subsection{Memory-Efficiency of Optimized Coding Matrices} 
\begin{figure}[tp]
    \centering
    \includegraphics[width=\linewidth]{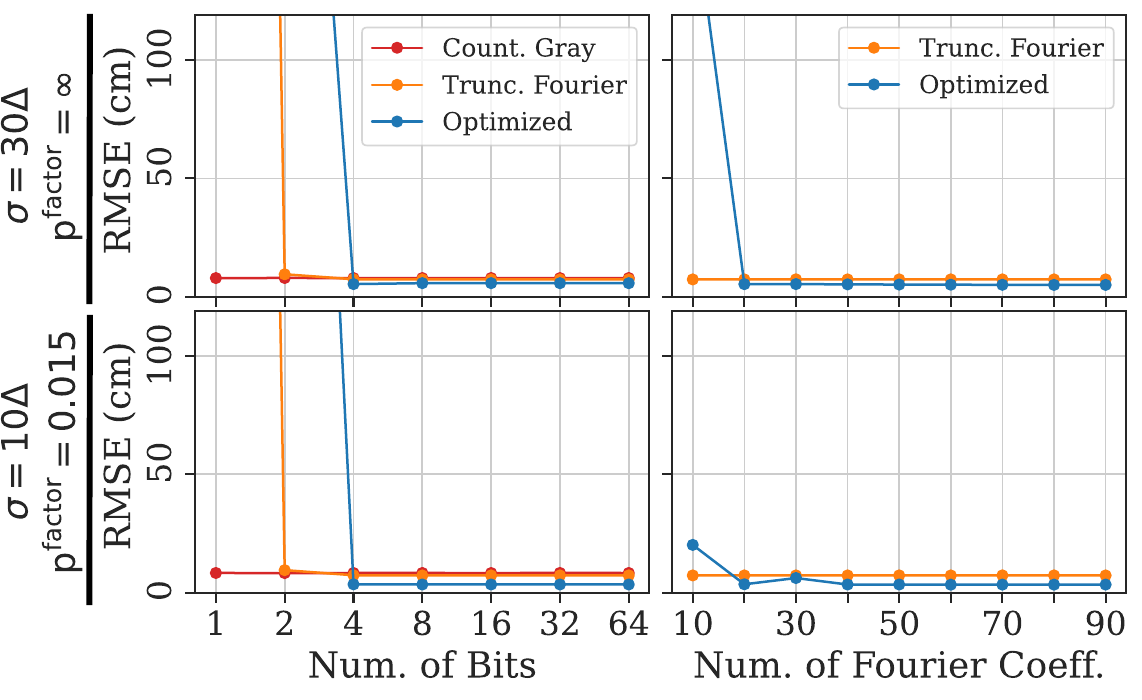}
    \caption{\textbf{Memory efficiency analysis of coding functions.} This figure plots the RMSE of each compressive code as a function of bit depth (left col) and number of Fourier coefficients (right col) used to represent the coding matrix, evaluated at low SBR ($0.1$). The top row presents results for band-limited coding functions with infinite peak power, while the bottom row shows results for the peak power-limited coding functions.}
    \label{fig:bit_depth_fourier}
\end{figure}

One benefit of Fourier and Gray coding functions is that they have memory-efficient representations~\cite{ardelean2023spad, Gutierrez-Barragan_2023_ICCV}. A naive in-pixel approach would require storing the full $64$-bit coding matrix, which, unless shared by multiple pixels, would lead to a sizeable in-sensor memory overhead. Therefore, an efficient representation of the optimized coding matrix is essential for on-chip implementation. To assess this, we evaluate the performance (Fig.~\ref{fig:bit_depth_fourier}) of the optimized coding functions when quantized to values between 1 and 64 bits and reconstructed using 10 to 90 Fourier coefficients. We find that both peak power-limited and band-limited coding matrices can be represented using as few as $4$ bits or $40$ Fourier coefficients with minimal performance loss. Gutierrez-Barragan et al.~\cite{Gutierrez-Barragan_2023_ICCV} recently showed that $8$-bit Fourier codes, when implemented on the UltraPhase chip~\cite{ardelean2023spad}, consume less power than data-intensive codes like coarse histogramming~\cite{gyongy2020highspeed}. Together, these findings suggest that our proposed coding functions have a feasible on-chip implementation under typical in-sensor memory constraints. Please see the supplement for memory analysis of additional coding functions.

\section{Discussion and Limitations}\label{sec:conculation}
Hardware constraints, such as laser power, sensor data rates, system bandwidth, and the illumination waveform, strongly influence the performance and practicality of single-photon 3D cameras. While compressive histograms were introduced to reduce data rates, they perform suboptimally under real-world hardware limitations. We propose a constrained optimization approach that identifies a set of high-performing coding functions while adhering to hardware constraints. We show that our method consistently outperforms compressive histograms across a wide range of scenarios. We hope this work takes a step toward enabling low-cost, low-power, and practical single-photon 3D imaging systems.

\smallskip
\noindent \textbf{Poisson noise in training neural networks:} Poisson noise is not differentiable. Therefore, to train our illumination, we approximated Poisson noise through Gaussian noise where the mean equals the variance. However, at low photon counts, this approximation is weak. In the photon-starved regime, our optimized coding functions can be prone to large depth errors, likely due to this approximation. A possible workaround is to optimize the illumination with Gaussian noise and then optimize the coding matrix using Poisson noise to improve the code's ability to adapt to single-photon sensors at low SNR.

\smallskip
\noindent \textbf{Scene Dependent Considerations:} We proposed a framework for designing coding functions optimized to a specific hardware impulse response. However, our optimization did not account for scene-dependent factors such as indirect reflections, surface albedo, or geometry. While the experimental and simulation results show that several coding functions are fairly robust to indirect reflections, this was not explicitly optimized. On the other hand, the performance of the peak power-limited coding functions may degrade due to the presence of multiple peaks in the illumination waveform. A possible solution would be to incorporate scene response into our network by adding a convolution layer.

\smallskip
\noindent \textbf{Framework for practical time-of-flight imaging:} While this paper focused on finding optimal coding functions for single photon ToF, this optimization framework is potentially viable for other imaging modalities that encode a transmitted signal, provided the decoding is differentiable. Examples include correlation-based ToF (C-ToF) 3D imaging, NLoS imaging, FLIM imaging, and gated imaging. Consider a C-ToF camera whose illumination is temporally modulated according to the function $m(t)$. A brightness value is computed as the correlation between the radiance incidence $r(t)$ and the sensor exposure function $d(t)$. Our proposed framework can be applied to finding high-performing modulation $m(t)$ and demodulation functions $d(t)$ that adhere to hardware constraints. Furthermore, our model has the potential to be extended to other ToF imaging methods such as NLoS and FLIM. However, we anticipate that applications needing full temporal information will achieve lower compression due to increased signal complexity and photon noise.


\bibliographystyle{IEEEtran}
\bibliography{11_references}


\begin{IEEEbiography}[{\includegraphics[width=1in,height=1.25in,clip,keepaspectratio]{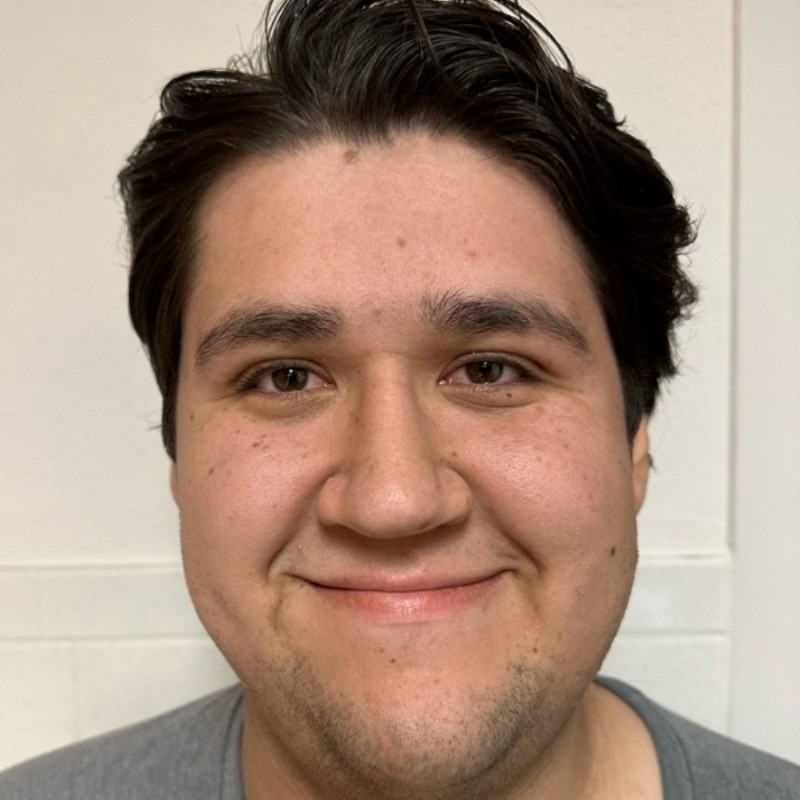}}]{David Parra}
is a third-year PhD student in Computer Science at the University of Wisconsin-Madison, advised by Prof. Andreas Velten. His research interests are in computational imaging, computer vision, and deep learning. He earned his Bachelor's degree from the University of Arizona (2021). 
\end{IEEEbiography}

\begin{IEEEbiography}[{\includegraphics[width=1in,height=1.25in,clip,keepaspectratio]{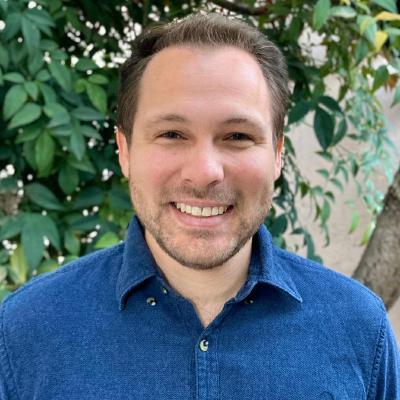}}]{Felipe Gutierrez-Barragan}
is a Senior Software Engineer at Ubicept, developing efficient image processing algorithms optimized for computer vision systems and emerging sensors such as high-speed HDR SPAD cameras. He received his B.S (2016), M.S. (2019), and Ph.D. (2022) in Computer Sciences from the University of Wisconsin-Madison.  His Ph.D. research focused on practical modifications to spad-based and indirect time-of-flight 3D cameras, aiming to reduce their power consumption and data bandwidth while preserving or enhancing their precision. His research interests are in computational imaging, computer vision, and machine learning.
\end{IEEEbiography}

\begin{IEEEbiography}[{\includegraphics[width=1in,height=1.25in,clip,keepaspectratio]{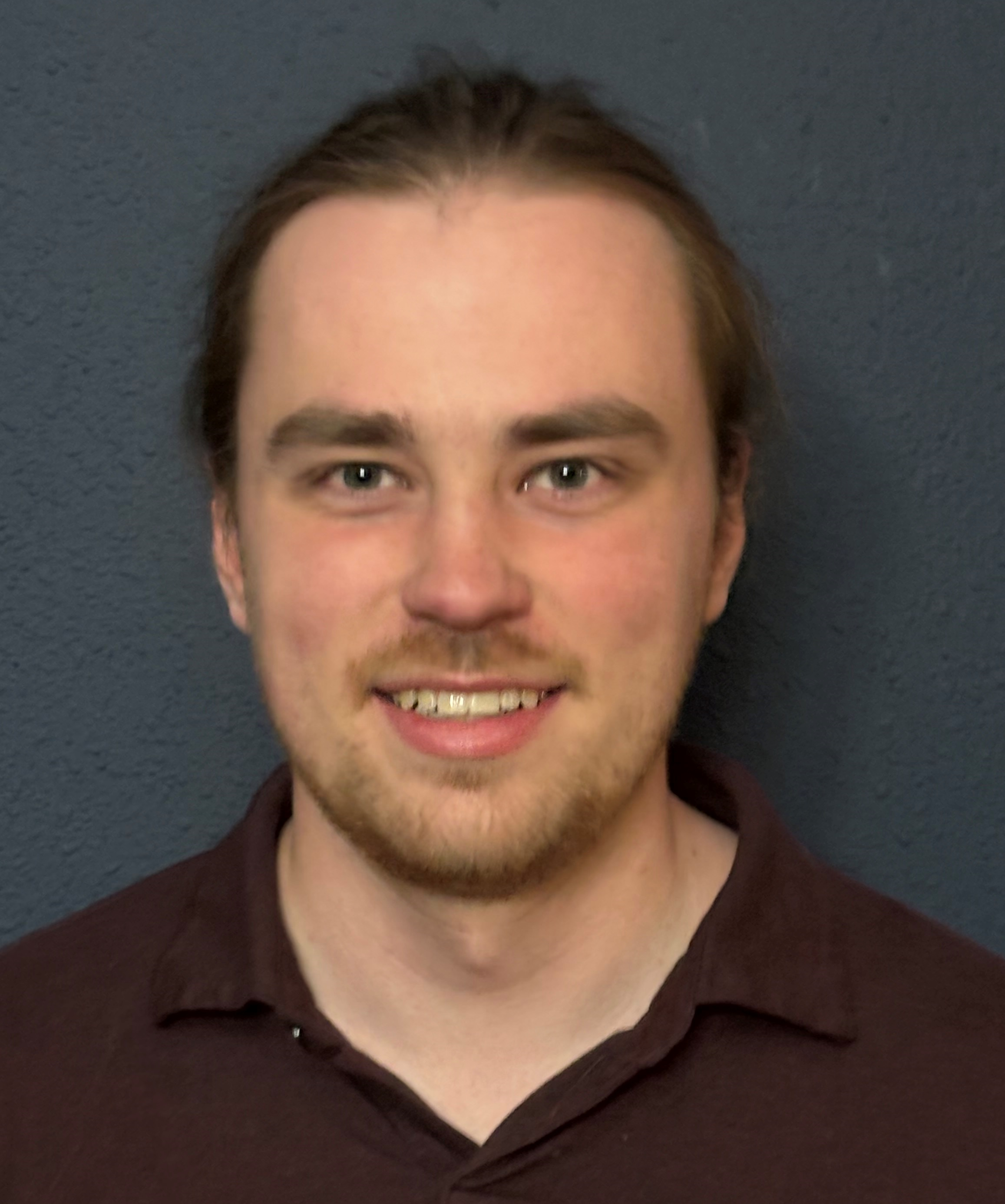}}]{Trevor Seets}
received a Ph.D. (2025) in Electrical Engineering from University of Wisconsin-Madison, Madison, WI in 2025. Currently, Trevor works as a Senior Software Engineer at Ubicept. His research interests include machine learning, statistical signal processing, and computational imaging.
\end{IEEEbiography}

\begin{IEEEbiography}[{\includegraphics[width=1in,height=1.25in,clip,keepaspectratio]{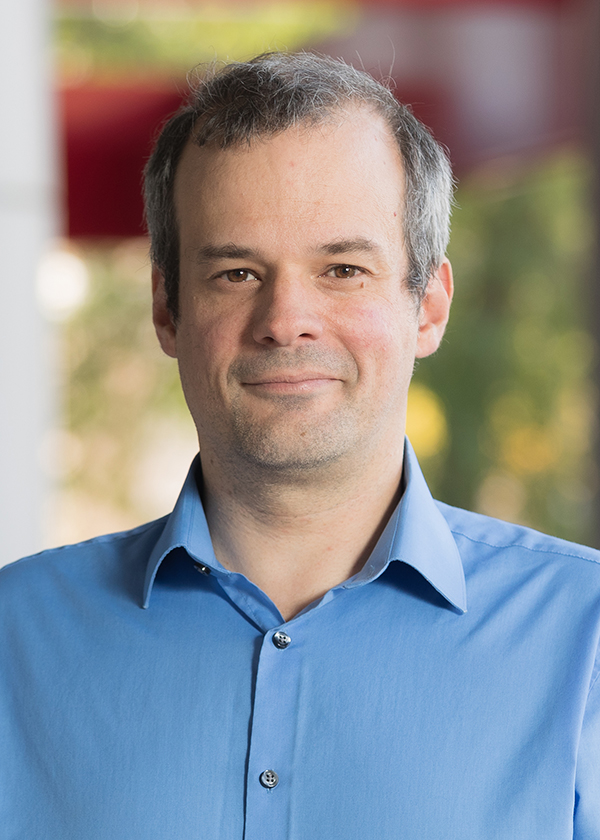}}]{Andreas Velten}
is Associate Professor at the Department of Biostatistics and Medical Informatics and the Department of Electrical and Computer Engineering at the University of Wisconsin-Madison and directs the Computational Optics Group. He obtained his PhD in Physics at the University of New Mexico in Albuquerque and was a postdoctoral associate of the Camera Culture Group at the MIT Media Lab. He has been included in the MIT TR35 list of the world's top innovators under the age of 35 and is a senior member of the National Academy of Inventors and SPIE as well as a fellow of Optica. He is co-Founder of Onlume, a company that develops surgical imaging systems, and Ubicept, a company developing single photon imaging solutions.
\end{IEEEbiography}

\ifpeerreview \else





\fi

\end{document}